\documentclass[final,12pt]{elsarticle}

\usepackage{graphicx}  
\usepackage{comment}
\usepackage{subfig}
\usepackage{amsfonts,amsmath,dsfont}
\usepackage{booktabs}
\usepackage{multirow}
\usepackage{url}
\usepackage{array}
\usepackage{xcolor}
\usepackage{graphics}
\usepackage[utf8]{inputenc}

\newcommand{\change}[1]{#1}
\newcommand{\Depression}{\textit{Depression}}
\newcommand{\Bipolar}{\textit{Bipolar}}
\newcommand{\Anxiety}{\textit{Anxiety}}
\newcommand{\SuicideWatch}{\textit{SuicideWatch}}

\journal{Future Generation Computer Systems}

\begin{document}

\begin{frontmatter}



\title{Predicting User Emotional Tone in Mental Disorder Online Communities\footnote{Citation format: Bárbara Silveira, Henrique S. Silva, Fabricio Murai, Ana Paula C. da Silva,
Predicting user emotional tone in mental disorder online communities,
Future Generation Computer Systems,
Volume 125,
2021,
Pages 641-651,
ISSN 0167-739X,
https://doi.org/10.1016/j.future.2021.07.014 \cite{silveira2021fgcs}.}}


\author{Bárbara Silveira}
\author{Henrique S.\ Silva}
\author{Fabricio Murai}
\author{Ana Paula C.\ da Silva}

\affiliation{organization={Department of Computer Science, Universidade Federal de Minas Gerais},
            country={Brazil}}



\begin{abstract}
\change{In recent years,} Online Social Networks \change{have become} an important medium \change{for people} who suffer from mental disorders \change{to share} moments of hardship\change{,} and receive emotional and informational support. \change{In this work,} we analyze how discussions in Reddit communities related to mental disorders can help improve the health conditions of their users. Using \change{the} emotional tone of users' \change{writing} as a proxy for emotional state, we uncover relationships between \change{user interactions and state changes}. First, we observe that authors of negative posts often write rosier comments after engaging in discussions, indicating that users' emotional state can improve due to social support. 
Second, we build models based on SOTA text embedding techniques and RNNs to predict shifts in emotional tone. This differs from \change{most of related work}, which focuses primarily on detecting mental disorders from user activity. We \change{demonstrate the feasibility of accurately predicting} the users' reactions to the interactions experienced in these platforms, and present some examples which illustrate that the models are correctly capturing the effects of comments \change{on the author's} emotional tone. Our models \change{hold promising implications for interventions to} provide support \change{for people struggling with mental illnesses}. 
\end{abstract}



\begin{keyword}
Emotional tone \sep Sentiment analysis \sep Mental health disorders \sep Reddit \sep Online communities \sep Emotional tone prediction
\end{keyword}

\end{frontmatter}


\section{Introduction}

 In recent years, statistics from the World Health Organization (WHO) \change{show a marked} increase in the total number of people \change{globally} that suffer \change{from mental disorders}; 1 in 4 people are affected by these illnesses during their lives~\cite{world2013mental}. WHO's \change{latest} report indicates \change{that depression cases have} grown 18\% between 2005 and \change{2015, affecting 322 million people}. Anxiety, in turn, affects 264 million people and bipolar disorder impacts around 60 million people. In their most severe form, these disorders can lead to suicide\change{, approximately} 800 thousand people \change{commit} suicide \change{anually} \cite{world2017depression}.

Despite these alarming numbers, \change{many mental disorders go untreated}. WHO reports that 3 \change{out of} every 4 affected people do not \change{receive} any type of treatment\change{;} 45\% of the world population live in countries with less than 1 psychiatrist for every 100 thousand people.\footnote{\url{who.int/mental_health/evidence/atlas}} Moreover, health systems do not provide adequate treatment for such disorders. WHO shows that in low and medium income countries, between 76\% and 85\% of the people with mental disorders do not receive treatment, whereas this number ranges between 35\% and 50\% in high income countries.

The combination of scarce resources, social stigma associated with mental disorders~\cite{barney2006stigma} and reluctance to ask for help prevents most \change{affected people from receiving much-needed assistance}. To overcome these challenges, new ways of using communication systems have revolutionized the support offered to people who suffer from mental disorders, \change{through chat, call or email helplines}. More recently, Online Social Networks (OSNs) have been emerging as an important tool \change{for} modeling mental well-being, \change{as researchers are realizing} that people's online behavior can be analyzed for mass mental-health screening provided that privacy issues are adequately observed \cite{10.1371/journal.pone.0215476,facebook2018,mind2020,munmun2020}.

Originally focused \change{on} fostering friendships\change{, dating} and \change{on} image and video sharing, in \change{recent} years\change{, many OSNs -- and, in particular, Reddit -- also} started to connect people willing to share experiences related to mental disorders~\cite{Choudhury2014MentalHD,gkotsis2016language,gkotsis2017characterisation,park2018harnessing,thorstad2019,Saha2020.08.07.20170548,shen2017detecting,yusof2018assessing,murrieta2018depression,lee2018advanced,Gaur2018,coppersmith2018natural,wongkoblap2019modeling,fraga2018wi,yates2017depression}. Reddit is a forum website with social network features: it is composed \change{of} communities (subreddits) where users create and share content (including their experiences)\change{; ask} questions about a wide range of subjects.  A user can initiate a thread by publishing a post. The post can be replied by other users (or even by the post author) through comments. Comments, in turn, can be replied through other comments. 

In \change{this work, we aim} to analyze whether Reddit mental health disorders communities \change{help} improve the health conditions of their users\change{, and if so, how}.  Our main research questions are: 

\begin{quote}
\emph{Do online interactions in \change{reddit mental health} communities have a positive effect on the user emotional state\change{? If} so, can we model this effect and \change{accurately predict it?}}
\end{quote}

To \change{address these questions,} we analyze the changes in the emotional state of a user over a thread by using their emotional tone -- a proxy which can be computed \change{using} sentiment analysis tools. In other words, we analyze whether the \change{act of seeking help} in these networks is effective\change{, and} results in changes \change{in user expression}, once they are engaged on these communities. The main contributions of this article are:

\begin{enumerate}
    \item We characterize the relationship between emotional tones of post, comments and \change{the} author's last comment in our Reddit \change{dataset}. Our findings show that users typically reply \change{to initial posts with a} more positive tone \change{comments, which can}, in fact, alter the \change{emotional} state of those who started the discussion.
    \item We predict the reaction of users \change{in} mental disorder online communities to the interactions experienced in these platforms. We build \change{a recurrent neural network model} to predict shifts in the emotional tone of the users. \change{Our} prediction model \change{exhibit low error rates}, in spite of the inherent \change{task difficulty}. Although the tone of the author's last comment is strongly correlated with that of the comments in a thread, more extreme shifts in emotional tone depend on the messages' contents and the order in which they appear.
    \item We include \change{illustrative} examples of threads \change{where} there is improvement in the author's emotional tone, which is accurately predicted by our models. Our results, which rely exclusively on posting and behavioral history gathered from  social media websites, could provide supplementary data to be applied in clinical care, providing timely interventions and possibly reaching populations \change{that are} difficult to access through traditional clinical approaches. 
\item \change{We \change{release} all \change{our} code and scripts on a GitHub public repository\footnote{Code repository: \url{https://github.com/HenrySilvaCS/SentiMentalHealth}}}.
\end{enumerate}

In summary, we analyze the user’s sentiment shifts after interacting with \change{their} peers on Reddit. Our analysis\change{,} as well \change{as} the proposed models\change{,} could assist in interventions promoted by health care  professionals to provide support to people suffering from mental health illnesses. Moreover, \change{our work} is of significant contribution to the growing literature on  machine learning applications in mental health, \change{and holds} practical implications for early intervention and prevention of mental health disorder\change{s}. 

The rest of this paper is organized as follows. We first discuss \change{related work} in Section \ref{sec:related}. Next, in Section \ref{sec:methodology} we present our approach for modeling emotional tone shifts, \change{while} in Section \ref{sec:sequenciaAtividades} we characterize the relationship between emotional tones of post, comments\change{,} and author's last post in our Reddit data. We then describe the experimental setup used to evaluate the prediction model in Section \ref{sec:experimental}, while Section \ref{sec:evaluation} provides the \change{model evaluation}. \change{Finally}, in Section \ref{sec:discussion}, we discuss the \change{results,} present our conclusions \change{and discuss future work}.



\section{Related Work}
\label{sec:related}

\change{In-person consultations have been historically complemented by various tools to assist people who suffer from mental health disorders.} For example, \change{volunteer-based crisis hotlines are} widespread in many countries, such as Argentina\footnote{https://www.casbuenosaires.com.ar/ayuda}, Brasil\footnote{https://www.cvv.org.br/}, United States\footnote{https://www.crisistextline.org/texting-in}, France\footnote{http://www.suicide-ecoute.fr/}, among others.  Authors in~\cite{Althoff2016NaturalLP} analyze conversations that took place via SMS \change{with} individuals related to the \textit{Crisis Trends} organization.\footnote{https://crisistrends.org/} The study evaluated the \change{counselors' behavior} given that \change{their} crucial role in supporting the individuals that use the service. One of the main conclusions is that \change{the most effective} counselors are more aware of the trajectory of the conversation, reply to the messages in a more creative fashion, without using generic sentences\change{,} and \change{quickly identify} the focus of the problem that the individual has, \change{contributing} to its solution. \change{In contrast, we use machine learning models to automatically capture characteristics of the interactions that contribute to improving users' emotional state in OSNs.}

\change{The} pervasiveness of OSNs in our daily lives \change{has revolutionized} the support offered to people who suffer from some type of mental disorder, pushing the medical community to understand how new technologies can be effective in identifying users who suffer from these illnesses and, at the same time, 
to measure the effectiveness of online interventions and the
role played by social support in online communities.
For instance, a set of works address the detection of anxiety, \change{depression, bipolar depression} and suicide \change{risk} in online social networks \cite{shen2017detecting,yusof2018assessing,murrieta2018depression,lee2018advanced,chen2018mood,wolohan-etal-2018-detecting,Gaur2018,dutta2018measuring,coppersmith2018natural,wongkoblap2019modeling,gruda2019feeling,sahota2019bipolar,baba2019detecting,Saha2020.08.07.20170548}, either through data characterization or machine learning algorithms for \change{classification. By understanding, modeling and predicting} the behavior of OSN users, \change{these works can help} propose policies that contribute to \change{improving the live of those} affected by these problems. 


Among \change{mental health studies using} Reddit data, many papers address the problem of predicting mental health \change{status} through machine learning methods, such as Support Vector Machine (SVM) \cite{shing-etal-2018-expert,10.1145/3134727,pirina-coltekin-2018-identifying,info:doi/10.2196/jmir.9840}, Logistic Regression \cite{10.1145/3173574.3174240,ireland-iserman-2018-within,de2016discovering} and Deep Learning \cite{shen2017detecting,gkotsis2017,yates2017depression,10.1145/3159652.3159725,ive-etal-2018-hierarchical}. \change{Those} papers framed their research questions as
\change{classification tasks, in contrast to our approach of predicting the effect of the social interactions as a regression task}. For instance, authors in \cite{ive-etal-2018-hierarchical} applied a hierarchical Recurrent Neural Network (RNN) architecture to the classification of posts related to mental health. The work of \cite{10.1145/3134727} uses SVM for inferring stress expression in Reddit posts. Authors in \cite{pirina-coltekin-2018-identifying} present a set of classification experiments for identifying depression in posts gathered from social media platforms. In \cite{gkotsis2017characterisation}, the authors analyze posts from the social media platform Reddit and develop classifiers to recognize and classify posts related to mental illness according to 11 disorder themes.




\change{All} the aforementioned studies focus on detecting people who suffer from some mental illnesses\change{, classifying} the type of their disorder, \change{and} analyzing how they behave and express themselves on various online social networks. 
Our work differs from those since we model \change{a user's} ``state'' during the interval in which \change{they interact} in a thread to predict changes \change{in emotional tone}, which is, to our knowledge, the first attempt of the kind. Monitoring \change{user activities} and their interactions to predict \change{granular} emotional ups and downs \change{affords} more effective \change{and timely} interventions.

\change{Characterizing an individual's emotional state (e.g., sadness, happiness, anger, fear) can be useful in understanding psychological processes from mental health and depression \cite{coppersmith-etal-2014-quantifying,edwards2017}, \change{since} emotions and behavior may fluctuate rapidly as an individual interacts with the environment \cite{loveys-etal-2017-small}. However, measuring or assessing emotions is not always straightforward and there is a body of works that aims at providing some guidelines for emotions measurement \cite {mauss2009}. 
Here, the emotional state of a user is measured by the emotional tone (EmT) \cite{kayla2018}, which is the balance between positive and negative emotion expressed in their messages.}

\section{Modeling Shifts in Emotional Tone}
\label{sec:methodology}


\change{In what follows, we conduct an analysis which reveals interesting} relationships between \change{a user's interactions and emotion changes} in a subreddit.



\noindent \textbf{Emotional tone.} Following a body of \change{related work}, we measure the intensity of emotion by means of sentiment scoring mechanisms \cite{loveys-etal-2017-small,chancellor2016,DBLP:conf/cscw/ChoudhuryCHH14}. \change{From existing} sentiment analysis tools, we \change{choose VADER~\cite{hutto2014vader}, a} sentiment analysis tool based on rules and on a lexical dictionary built specifically from \change{OSN} data. 
VADER computes four variables from a given text: positive, negative, neutral and compound. 
Compound is the combination of three first lexical variables normalized between -1 (extremely negative) and +1 (extremely positive) and, hence, is the \change{scalar value we assign as the} emotional tone \change{of a publication (i.e., post or comment)}.

\begin{figure*}[t!]
\centering
\includegraphics[width=\textwidth]{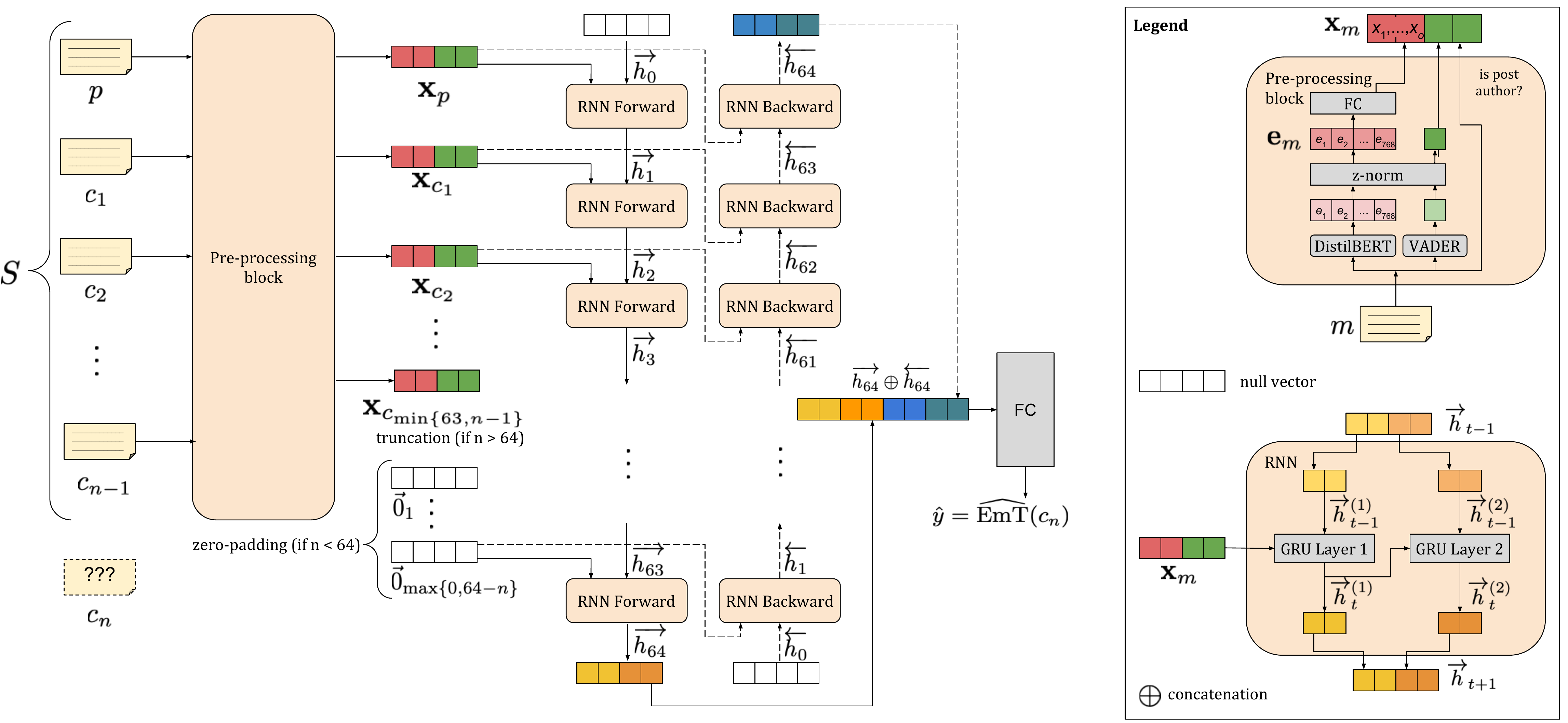}
\caption{Model architecture.}
\label{fig:architecture}
\end{figure*}

\noindent \textbf{Prediction task.} Let $p$ be the \change{(opening)} post \change{(often referred to as OP)}, written by author $u$, which initiates a thread. \change{Within the thread, post} $p$ is followed by chronologically ordered comments \change{ $c_1, c_2, \ldots$. It will be useful for us to denote by $c_n$ the} last comment made by $u$ in that thread. Define $\textrm{EmT}(.)$ as the emotional tone of a post or comment. Given the sequence $S=(p, \ldots, c_{n-1})$, we aim to predict the emotional tone $\textrm{EmT}(c_n)$ of the last comment made by the thread author. In other words, we want to infer the effect on the user emotional state prompted by the comments in $S$, \change{i.e..} we aim to measure the impact that emotional and informational support from these communities may provide to their members. Figure~\ref{fig:architecture} shows a diagram illustrating \change{the model proposed to address this task}. Posts and comments are fed as input to models that predict shifts in emotional tone.

\noindent \textbf{Model overview.} The input is the sequence $S=(p,c_1,\ldots,c_{n-1})$ of messages consisting of post $p$ and comments $c_1\ldots,c_{n-1}$ from a thread, sorted in chronological order, \change{without the} last comment made by the post author \change{($c_n$)}. Each message $m \in S$ is passed to a pre-processing block that extracts (i) embeddings using DistilBERT \change{\cite{sanh2020distilbert}}, (ii) emotional tone using \change{VADER,} and (iii) a variable indicating whether $m$ was written by the post author. In that block, the only trainable component is a Fully Connected (FC) layer. Embeddings are then forwarded through a Gated  Recurrent  Unit (GRU\footnote{We opted for GRUs because their performance is on par with LSTMs, but are computationally more efficient as they do not require a memory unit in addition to the state unit \cite{chung2014arxiv,jozefowicz2015pmlr}. Experiments with vanilla attention models did not improve the results \change{further}.}) based recurrent neural network (RNN) that can be either unidirectional or bidirectional and can contain either 1 or 2 layers (2-layer bidirectional RNN is shown). Hidden states are concatenated and passed through a second FC layer to output the prediction $\hat y = \widehat{\textrm{EmT}}(c_n)$ for the emotional tone of $c_n$. GRUs and second FC layer are also trainable.

\noindent \textbf{Limitations.} While we use emotional tone as a proxy for emotional state, it is not possible to evaluate whether there was a real improvement (or degradation) for the user due to the anonymity inherent to this OSN. Consequently, the metrics we use do not consider complex external aspects of the \change{user's life} which could help us make this assessment.
\change{Also, we cannot identify activities other than post/comment writing that the user may have performed on other threads, such as reading and down/up-voting.}
Another valid concern is sentiment score misratings, which are more likely to happen when the text contains ambivalent tones. For instance, the text ``Thank you. I am much worse now'' presents a clear negative emotional tone, but due to the etiquette pattern ``Thank you'', sentiment analysis tools like VADER might incorrectly infer an overall positive emotional tone. Since these are limitations intrinsic to such tools, we conducted  a separate analysis on the impact of etiquette words, \change{e.g., ``Thanks'', ``Thank you'', ``Thx''}, etc, in order to determine whether the model predictions could be severely affected by these types of misratings. After filtering out a set of predetermined etiquette words, we found that, for this specific dataset, fluctuations between the emotional tone of the original posts/comments and the emotional tone of the filtered ones were not significant. This is the case because: (i) there is a small fraction of posts/comments containing such etiquette patterns, and (ii) for the ones that did, removing the patterns did not change much the emotional tone. Therefore, we use the unfiltered data, since it makes the preprocessing simpler. For conciseness, the additional study is not included here.

\noindent \textbf{Pre-processing.} To train machine learning models for this task, we need to extract textual features from each publication. After experimenting with LIWC (Linguistic Inquiry and Word Count) features \cite{pennebaker01} and word embedding methods, we settled on using the latter. Particularly, we use Transformer architectures based on Deep Neural Networks, which achieve state-of-the-art performance in many NLP tasks. Since we envision the prediction task possibly as part of real time applications, we choose the pre-trained DistilBERT architecture, which is much lighter than BERT, the most popular variant\change{~\cite{bert2018}}. We do not fine-tune DistilBERT parameters to our data, as we focus our efforts on designing and training networks that can use these embeddings as input to predict shifts in \change{emotional tone}.

Each \change{message $m\in S$} is forwarded through DistilBERT to yield an embedding $\mathbf{e}_m \in \mathbb{R}^{768}$. We then concatenate two hand-engineered features to $\mathbf{e}_m$: the emotional tone $\textrm{EmT}(m)$\change{, and} a variable indicating whether $m$'s author is the post author. Together, these two features can provide emotional tone ``snapshots'' to the model. 
To avoid under-representing the hand-engineered features, we reduce the embedding dimensionality by passing it through a fully connected (FC) layer whose output size $o$ \change{is either 2, 14 or 62}, so that the resulting feature vector $\mathbf{x}_m$ size is \change{respectively 4, 16 or 64}:
\begin{equation}
    \mathbf{x}_m = [\textrm{FC}(\mathbf{e}_m),\, \textrm{EmT}(m),\, \textit{is\_post\_author}(m)] \in \mathbb{R}^{o+2}.
\end{equation}

After pre-processing, each sequence $S$ is represented by a matrix $\mathbf{X} = [\mathbf{x}_p; \mathbf{x}_{c_1};\ldots,\mathbf{x}_{c_{n-1}}] \in \mathbb{R}^{n \times (o+2)}$ and the scalar $y = \textrm{EmT}(c_n)$. Since the number of comments varies across threads, we truncate longer matrices (0.4\% of all threads, no more than 0.9\% of any subreddit) to the first $64$ rows and zero-pad shorter matrices, so that every matrix is $64 \times (o+2)$. We use these representations to learn regression models and, in turn, to make predictions. Specifically, we choose to work with recurrent neural networks as a natural choice for processing sequences. These models will be described in detail in the Experimental Setup section.

\noindent \textbf{Datasets.} We gathered all posts, comments and metadata created between 2011 and 2017 from the four mental health related Reddit communities with the largest number of publications\footnote{Data is 
publicly available at \url{http://files.pushshift.io/reddit}.}, namely \change{\Depression, \SuicideWatch, \Anxiety\ and \Bipolar}~\cite{gkotsis2017characterisation}.

We focus this study \change{on the 2017 data, since during that year,} there was an \change{unprecedented} volume of user interactions on these subreddits, which is important for training and testing our prediction models. We present some descriptive statistics of the dataset in Table \ref{tab:statistics}. \Depression\ is the largest community in number of unique users\change{, while} users in \Bipolar\ are the most active, writing at least twice as often than users in other communities, which shows a stronger engagement in that subreddit. 

\begin{table*}[t]
\setlength{\tabcolsep}{2.5pt}
\centering
\caption{Statistics of interactions on each subreddit. $^\dagger$Threads used in the prediction task have to satisfy some constraints described later, in Section~\ref{sec:sequenciaAtividades}.}
\scriptsize
\begin{tabular}{l|rrrrrrrrrr}
\toprule
& \multicolumn{8}{c}{\textbf{Subreddit}} & \\
\midrule
 { }&
  \multicolumn{2}{c}{\textbf{Anxiety}} &
  \multicolumn{2}{c}{\textbf{Bipolar}} &
    \multicolumn{2}{c}{\textbf{Depression}} &
  \multicolumn{2}{c}{\textbf{SuicideWatch}} &
  \multicolumn{2}{c}{\textbf{Total}} \\
 \midrule 
 \textbf{Publications (numbers)}  & \multicolumn{2}{r}{} & \multicolumn{2}{r}{} & \multicolumn{2}{r}{} & \multicolumn{2}{r}{} & \multicolumn{2}{r}{}\\
{Threads (posts)}            & \multicolumn{2}{r}{25,574}  & \multicolumn{2}{r}{15,825} & \multicolumn{2}{r}{70,950}  & \multicolumn{2}{r}{29,486}  & \multicolumn{2}{r}{141,835}  \\ 
{Comments}        & \multicolumn{2}{r}{150,155} & \multicolumn{2}{r}{136,824} & \multicolumn{2}{r}{448,005} & \multicolumn{2}{r}{182,731} & \multicolumn{2}{r}{917,715} \\
{Threads used in prediction task$^\dagger$ }            & \multicolumn{2}{r}{11,956}  & \multicolumn{2}{r}{8,165} & \multicolumn{2}{r}{30,039}  & \multicolumn{2}{r}{12,957}  & \multicolumn{2}{r}{63,117}  \\ 
\midrule
\textbf{Users (numbers)}  & \multicolumn{2}{r}{} & \multicolumn{2}{r}{} & \multicolumn{2}{r}{} & \multicolumn{2}{r}{} & \multicolumn{2}{r}{}\\
{Unique users}     & \multicolumn{2}{r}{36,616}  & \multicolumn{2}{r}{11,363} & \multicolumn{2}{r}{85,706}  & \multicolumn{2}{r}{38,121}  & \multicolumn{2}{r}{154,114} \\
{Users who post (posting users)}   & \multicolumn{2}{r}{16,394}  & \multicolumn{2}{r}{6,043}  & \multicolumn{2}{r}{41,813}  & \multicolumn{2}{r}{20,844}   & \multicolumn{2}{r}{78,499} \\
{Users who comment (commenters)}   & \multicolumn{2}{r}{31,783}  & \multicolumn{2}{r}{10,190}  & \multicolumn{2}{r}{72,268}  & \multicolumn{2}{r}{30,113}   & \multicolumn{2}{r}{129,908} \\

\midrule 
\textbf{Others (median and max.)}  & \multicolumn{2}{r}{} & \multicolumn{2}{r}{} & \multicolumn{2}{r}{} & \multicolumn{2}{r}{} & \multicolumn{2}{r}{}\\ 
%
{Posts per posting user}   & {1} & {92} & {1} & {247} & {1} & {99} & {1} & {49} & {1} & {247}\\
{Comments per commenter}   &  {2} & {1,665} & {3} & {2,326} & {2} & {1,850} & {2} & {3,999} & {2} & {4,018} \\
{Comments in thread }   &  {3} & {241} & {6} & {134} & {3} & {4,938} & {3} & {284} & {3} & {4,938}\\
{Post length (in chars)}   &  {692} & {24,952} & {576} & {28,840} & {594} & {39,863} & {715} & {36,683} & {634} & {39,863} \\
{Comment length (in chars)}   &  {213} & {9,791} & {188} & {9,796} & {148} & {9,997} & {146} & {9,947} & {164} & {9,997} \\
\bottomrule
\end{tabular}
\label{tab:statistics}
\end{table*}

\section{Characterization of Emotional Tone Shifts}
\label{sec:sequenciaAtividades}

\begin{figure}[t!]
\centering
\includegraphics[width=0.8\columnwidth]{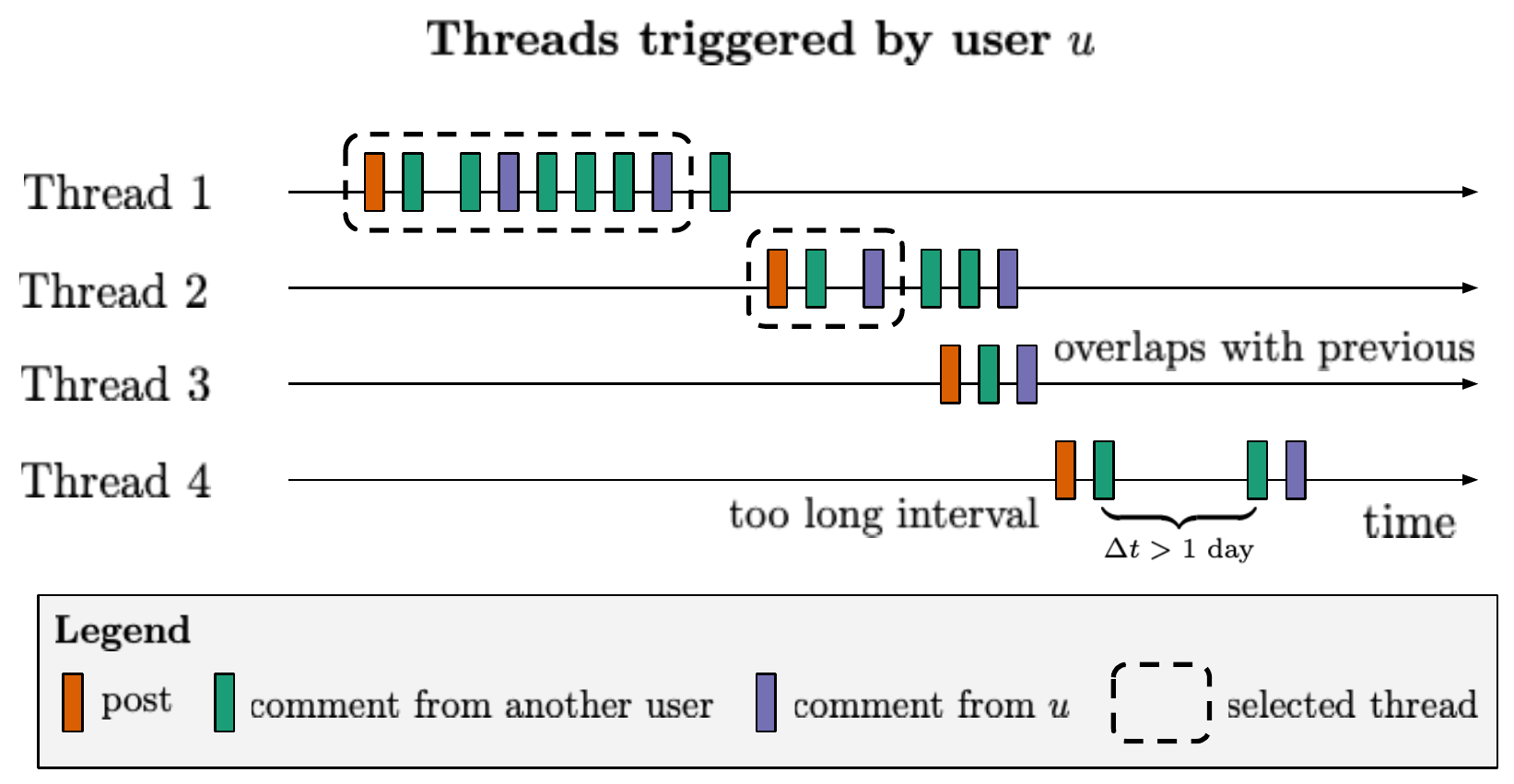}
\caption{Toy example of the thread selection process. Threads must include 1 or more interactions with other users followed by the author's response, must not overlap with other threads or have long intervals between interactions.}
\label{fig:threads}
\end{figure}

To study the effect of \change{user interactions} in a thread, we eliminate obvious confounders, such as simultaneous participation in multiple threads. To accomplish that, we select, for each user, the segments of threads started by \change{them} which:
\begin{itemize}
    \item begin with a post and end with \change{their} last comment prior to becoming active in another thread\change{,}
    \item  contain at least one comment from another user\change{, \textbf{and}}
    \item  have less than 24 hours between consecutive publications.
\end{itemize}
We set the 24 hour limit to control \change{for the effect that longer periods} of time might have on the user emotional state. Figure \ref{fig:threads} shows a toy example of the thread selection process for a given user. All threads were triggered by the same user. In this example, \change{all messages in Thread 1 until the last comment from the post author are considered, whereas Thread 2 is truncated at the last comment from the post author prior to the overlap with Thread 3. Thread 3 is discarded due to the said overlap. Last,} Thread 4 is discarded because the interval between its 2$^\textrm{nd}$ and 3$^\textrm{rd}$ comments exceeds 24 hours.

\change{In what follows we investigate characteristics related to the emotional tone in this data.}

\begin{figure*}[t]
\centering
\captionsetup[subfloat]{position=top,labelformat=empty,farskip=0pt,captionskip=0pt}
\subfloat[][]{\includegraphics[trim=0 10 0 0,clip,width=0.95\textwidth]{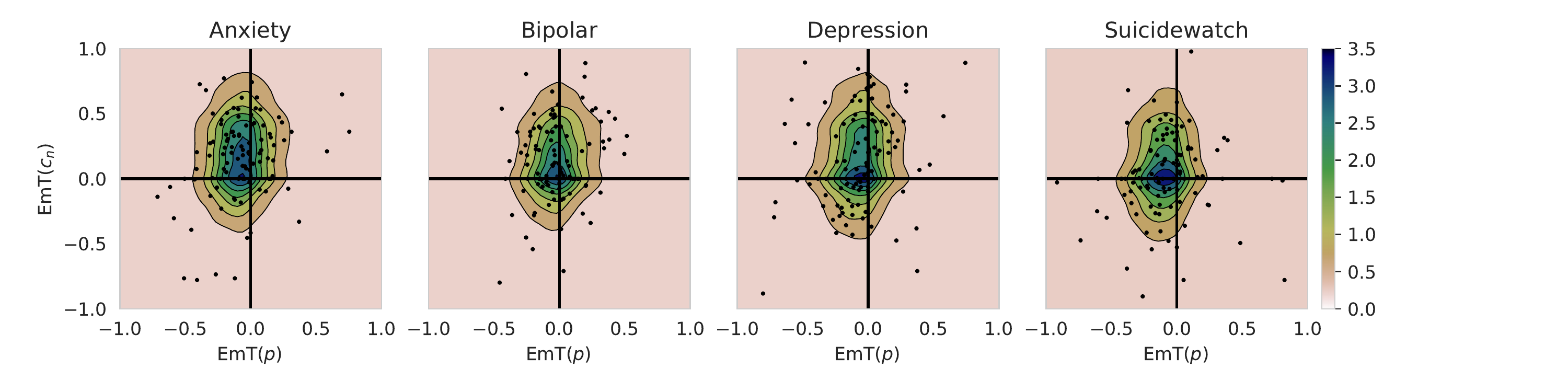}}\\
\subfloat[][]{\includegraphics[trim=0 10 0 22,clip,width=0.95\textwidth]{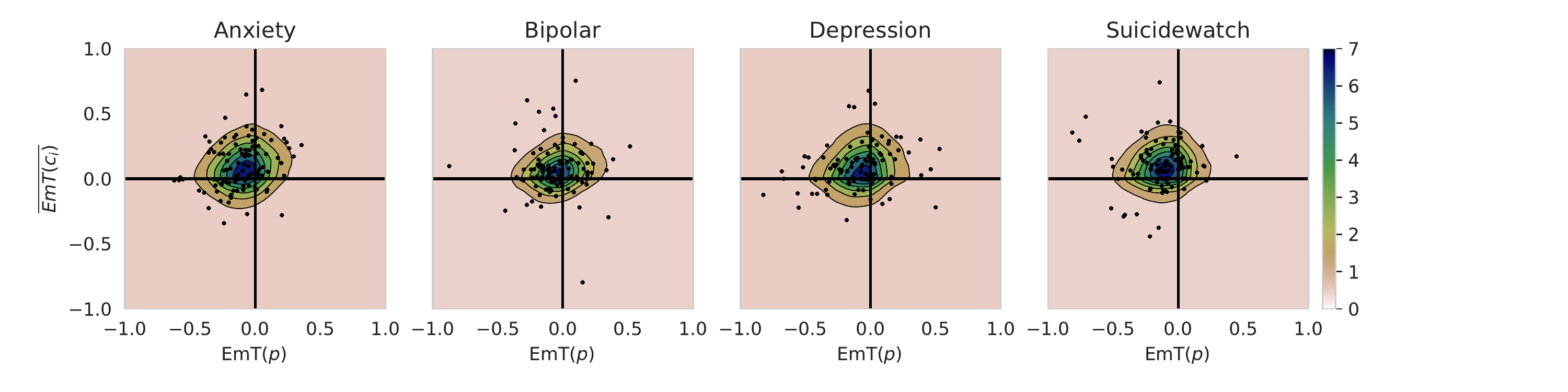}}\\
\subfloat[][]{\includegraphics[trim=0 10 0 22,clip,width=0.95\textwidth]{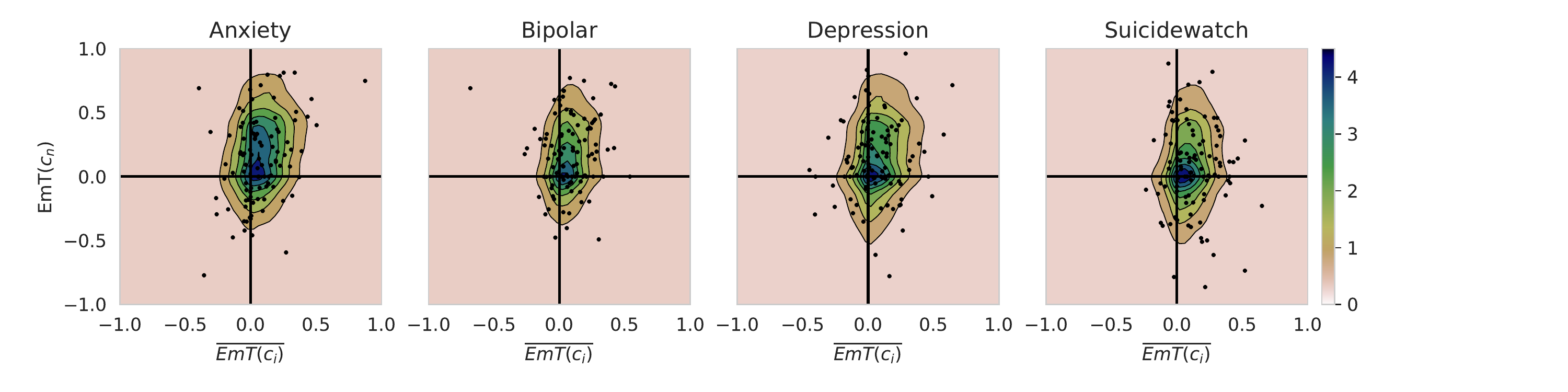}}
\caption{Pairwise joint distributions of emotional tone of post, $\textrm{EmT}(p)$, emotional tone of author's last comment, $\textrm{EmT}(c_n)$, and average emotional tone in thread, $\overline{\textrm{EmT}(c_i)}$, using a 2-D Gaussian kernel density estimator. Points in each plot represent 100 threads sampled uniformly at random.
\textbf{Top:} $\textrm{EmT}(p)$ vs.\ $\textrm{EmT}(c_n)$. 
\textbf{Middle:} $\textrm{EmT}(p)$ vs.\ $\overline{\textrm{EmT}(c_i)}$. 
\textbf{Bottom:} $\overline{\textrm{EmT}(c_i)}$ vs.\ $\textrm{EmT}(c_n)$. 
}
\label{fig:vader-hist2dCompPostComoUltimoComm}
\end{figure*}

\noindent \textbf{Emotional tone in threads.} We investigate how the emotional tone of (i) the posts, (ii) the last comment made by thread authors and (iii) the comments made by other users is distributed for each subreddit. In particular, instead of considering the set of all subreddit comments to study (iii), we take the average emotional tone in each thread to avoid \change{overweighting} longer threads. In our data, posts are more likely to express negative emotional tone,  but last comments are more likely to exhibit the opposite sentiment. The median $\textrm{EmT}$ of comments from others and that of last comments are higher than that of posts\change{, indicating that the commenters are generally attempting to be supportive and cheer up the thread author.}. Moreover, the 1st, 2nd and 3rd quartiles of the last comments' $\textrm{EmT}$ are higher than those of the posts, thus suggesting an improvement on the emotional tone of the thread authors. In what follows, we investigate this and other relationships in detail using the heatmaps show in Figure~\ref{fig:vader-hist2dCompPostComoUltimoComm}.


\noindent \textbf{Relationship between post and last comment's emotional tone.} We investigate shifts in emotional tone between the moment a user writes a post and the moment \change{they write} their last comment on the thread.
$\textrm{EmT}(p)$ has negative mean, while $\textrm{EmT}(c_n)$ has positive mean and larger variance than the former (Fig.~\ref{fig:vader-hist2dCompPostComoUltimoComm} (top)). High concentration in 2$^\textrm{nd}$ quadrant suggests emotional state improvement after interactions in a thread, i.e, authors of negative posts tend to write more positive comments at the end. In \Anxiety\ and \Depression, this tendency is even stronger. 
This variation corroborates the idea that comments written by other users may help those facing difficult situations. 
Moreover, low concentration in 4$^\textrm{th}$ quadrant indicates that decrease in emotional state is \change{uncommon (ranging from 6.3 to 10.2\% of the subreddit's threads)}. Not surprisingly, there \change{are few} points in the 3$^\textrm{rd}$ quadrant\change{,} which suggests that\change{, for some users,} there is hardly any emotional state improvement after interactions \change{in a thread.}

\noindent \textbf{Relationship between post and other comments' emotional tone.} 
The average emotional tone of comments in a thread has small positive mean and low variance (Fig.~\ref{fig:vader-hist2dCompPostComoUltimoComm} (middle)). High concentration in 2$^\textrm{nd}$ quadrant shows that comments are on average slightly more positive than the post that initiated the thread. 

\noindent \textbf{Relationship between last comment and other comments' emotional tone.}
Joint distribution  in Fig.~\ref{fig:vader-hist2dCompPostComoUltimoComm} (bottom) is similar to that of Fig.~\ref{fig:vader-hist2dCompPostComoUltimoComm} (top), but slightly shifted towards more positive \change{values in the $x$-axis}. 
Positive emotional tone in the last comment is correlated with positive emotional tone in other comments. Interestingly, we also observe an average of about 18\% of negative last comments, even when conditioning on a discussion thread that is positive on average. This suggests that\change{, for some users in these communities, conversational engagement is insufficient to improve} their emotional state.

Overall, our results indicate that being part of supportive online groups facing similar mental health disorders \change{can create some improvement in emotional state. Whether that improvement is temporary or creates a lasting impact is a subject for future research.} 

\section{Experimental Setup} \label{sec:experimental}

To leverage information from the sequence of interactions represented by \change{feature matrix $\mathbf{X}$ for predicting $\textrm{EmT}(c_n)$}, we build recurrent neural networks (RNNs) such as the one shown in Figure~\ref{fig:architecture}, using Gated Recurrent Units (GRUs). We test several architectures using grid search: FC output size $o \in \{2,14,62\}$, unidirectional and bidirectional, with 1 (without dropout) or 2 layers (dropout probability in $\{0.0,0.1,0.2,0.5\}$), accounting for 30 configurations (see details in~\ref{sec:tuning}). The hidden state of an RNN layer is half the size of $\mathbf{x}_m$. Since this is a regression task, we pass a hidden state through another FC layer to obtain a scalar prediction \change{$\widehat{y}$ for $\textrm{EmT}(c_n)$}. For unidirectional RNNs, we use the hidden state associated with the last comment in $S$. For bidirectional networks, we concatenate the last hidden states from the forward and backward passes.

\noindent \textbf{A task-specific loss function.} For regression tasks, mean squared error (MSE loss) and mean absolute error (L1 loss) are typical choices of loss function. \textit{We found that training the model with these functions collapses predictions around the average}, preventing us from accurately forecasting more extreme values of $\textrm{EmT}(c_n)$. This is likely due to the inherent difficulty of predicting the user response (a common theme in social sciences), in which case it is ``safer'' to predict average behavior. To address this issue, we propose a \textbf{Weighted L1 loss}.
Because the distribution of $y \in [-1,1]$ is a bell-like curve, we bin observations w.r.t.\ $y$ and weight the absolute error for $(\mathbf{X},y)$ proportionally to the reciprocal of the number of elements in its bin. We use 10 equal-length bins\footnote{\change{We also tested with 5 bins, obtaining similar results.}}.

\noindent \textbf{Train/validation/test sets.} We use \change{an} 80-10-10\% random split\change{, stratified by weight bins}. Since threads are created over time, it may seem more natural to split them in chronological order instead. Yet, we observed that this produces higher error rates due to covariate shift (distribution of train and test splits differ). Hence\change{,} we advocate that models should be trained on threads from all the different months to capture seasonal differences. 


\noindent \textbf{Training.} We consider both training a unique model from the entire dataset (\textsc{Model}) and training one per subreddit (\textsc{Model-subreddit}). We optimize model parameters using batch size 32 and optimizer Adam over 20 epochs, with early stopping after 3 epochs without improvements \change{in} the validation loss. Training for most configurations ends within 8 epochs. We set model parameters to those that yielded the minimum validation loss.

\noindent \textbf{Computational resources.} All experiments were run on free Google Colab instances (Intel(R) Xeon(R) CPU@2.30GHz, 4 threads, 26G memory). DistilBERT embeddings were computed on TPU nodes at $\approx 40$ subreddit threads/s. Model training was done on nodes equipped with a Nvidia K80 GPU. Each model took 2 to 11min to train (average $\approx 7$min). Most epochs took 20 to 50s (average $\approx 46$ s). The complete grid search took 207min.

\section{Prediction Evaluation}
\label{sec:evaluation}

\change{In this section, we evaluate the proposed model. First,} we introduce three baselines used to evaluate the proposed model. Next, we present qualitative results, showing how responses $y$ and their respective predictions are distributed, and quantitative results, using the Weighted L1 Losses as performance metrics, illustrated with some prediction examples. We then analyze the ability of the proposed model to predict large shifts in emotional tone. Last, we discuss important implications of this study and possible applications.

\subsection{Baselines}
\label{sec:baselines}

To the best of our knowledge, this is the first work proposed to address this prediction task. In the absence of existing approaches, we consider two simple, but surprisingly effective heuristics, and a powerful regression model:
\begin{itemize}
    \item \textsc{Mean}: outputs the average emotional tone in $S$:
    \[\hat y_\textsc{Mean} = 1/|S| \sum_{m \in S} \textrm{EmT}(m).\]
    \item \textsc{Last}: outputs the emotional tone of the last comment in $S$:
    \[\hat y_\textsc{Last} = \textrm{EmT}(c_{n-1}).\]
    \item \textsc{XGBoost} (XGB): a gradient boosting ensemble of decision trees that achieves state-of-the-art performance in many prediction tasks \cite{xgboost}. As thread length varies and XGB requires a fixed-size input, we use a strategy widely used to aggregate features extracted by neural networks: we apply Max-Pooling and Average-Pooling to the message representations (along the time dimension) and concatenate the resulting vectors. Each message $m$ in a thread $S=(p, c_1, \ldots, c_{n-1})$ is represented\footnote{\change{The distinction between message $m$'s representation $\widetilde{\mathbf{x}}_m$ (used for generating XGB's input $\widetilde{\mathbf{x}}$) and $\mathbf{x}_m$ (used as the RNN's input) is that the former uses the raw text embedding $\mathbf{e}_m$, whereas the latter uses a linearly transformed version, $\textrm{FC}(\mathbf{e}_m$).}} as $\widetilde{\mathbf{x}}_m = [\mathbf{e}_m,\, \textrm{EmT}(m),\, \textit{is\_post\_author}(m)] \in \mathbb{R}^{770}$. Hence,
    \begin{gather*}
    \widetilde{\mathbf{x}} = \textsc{AveragePooling}([\widetilde{\mathbf{x}}_m]_{m \in S})  \oplus \textsc{MaxPooling}([\widetilde{\mathbf{x}}_m]_{m \in S}),\\
    \hat y_\textsc{XGB} = \textsc{XGBpredict}(\widetilde{\mathbf{x}}) \qquad \textrm{for $\widetilde{\mathbf{x}} \in \mathbb{R}^{2 \times 770}$},
    \end{gather*}   
    where $\oplus$ denotes concatenation. This captures both the overall characteristics of the messages in the thread as well as salient features present in any of the messages. For objective functions, we considered either the MSE or the Weighted L1 Loss (gradients and Hessians were derived and implemented \change{based on the previously defined bins}). However, the latter objective resulted in constant predictions. Therefore, we focus on the results for the MSE objective.\footnote{Other than that, we follow the same methodology used with the proposed model: we minimize the objective function with early stopping, perform hyper-parameter tuning and present the results of the model that yielded the minimum validation loss (see details in~\ref{sec:tuning}).}

\end{itemize}
We tested many variants of these baselines that proved to be less accurate, e.g., considering only comments from the branch of the thread that includes the author's last comment and/or excluding author's comments from \textsc{Mean}. \change{We therefore exclude them from this analysis.}

\subsection{Results}

In the following figures/tables, we abbreviate \change{\Anxiety\ (ANX), \Bipolar\ (BIP), \Depression\ (DEP) and \SuicideWatch\ (SUI)} to avoid cluttering. We select, in each case, the model that achieved the lowest validation loss. \change{The difference in loss between the selected hyperparameter setting and the others is at most 17.1\%, suggesting some level of robustness across settings.}

\begin{figure}[t!]
\centering
\includegraphics[trim=0 60 50 30,clip,width=0.95\columnwidth]{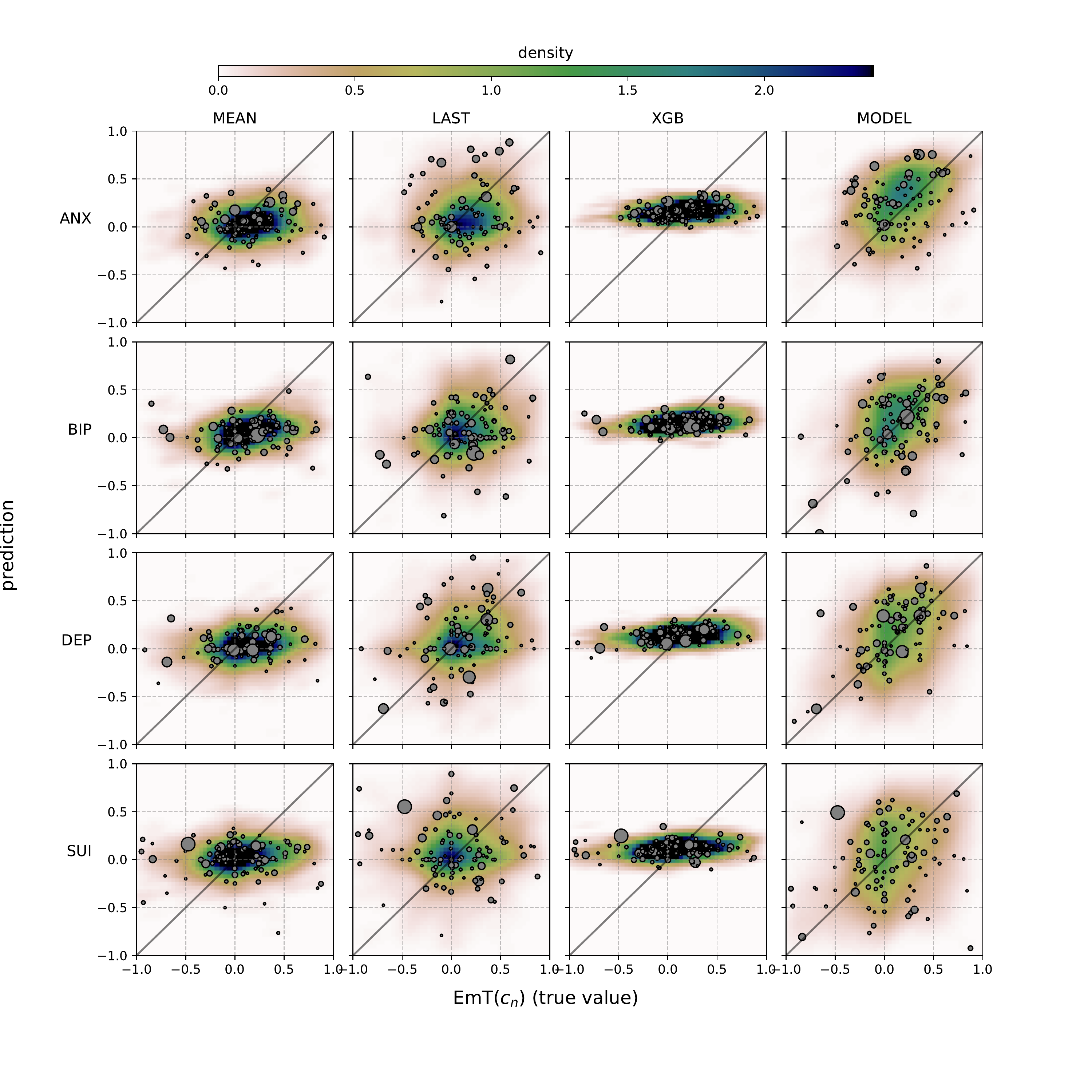}
\caption{Joint distribution of true and predicted values for \textsc{Mean}, \textsc{Last}, \textsc{XGB} and \textsc{Model} (color-code represents density) using a 2D Gaussian kernel density estimator. Each plot in a row includes 100 points illustrating predictions for the same randomly sampled threads (size represents number of comments; smallest is 2, largest is 64). \textsc{Mean} and \textsc{XGB} collapse predictions around average $y$. \textsc{Last} exhibits high density areas far from the diagonal. \textsc{Model} is best at predicting more extreme values.}
\label{fig:heatmap_bipolar}
\end{figure}

\noindent \textbf{Visual analysis.} The advantage of the proposed model over the baselines can be visualized from the heatmaps shown in Figure~\ref{fig:heatmap_bipolar}, which display the joint distribution of true and predicted values using a color-coded Gaussian kernel density estimator along with 100 random threads from the test set (circle sizes are proportional to number of comments; which vary from 2 to 64). We observe that \textsc{Mean}'s predictions are highly concentrated in $[-0.2,0.3]$. While \textsc{Last} spans predictions over a wider range, they are not well concentrated along the diagonal. Particularly, it results in large errors (vertical distance to diagonal) for some threads with more extreme -- especially negative -- responses (e.g., see leftmost points on BIP and SUI plots). \textsc{XGB} predictions are even more concentrated around the average than \textsc{Mean}. Although it yields the lowest L1 and MSE errors (see~\ref{sec:other_losses}), these predictions have no practical value. Conversely, \textsc{Model} exhibits predictions spread along the diagonal, showing a good correspondence with the true values.

\begin{table*}[ht!]
\setlength{\tabcolsep}{6pt}
\centering
\caption{Results w.r.t.\ the Weighted L1 Loss, which gives more weight to more extreme responses. \textsc{Model} is trained on entire 2017 dataset whereas \textsc{Model-Subreddit} is trained on target subreddit (column). \textsc{Model} outperforms all the baselines by at least $12.9\%$.}
\label{tab:test_loss}
\scriptsize
\begin{tabular}{lrrrr}
\toprule
\multirow{2}{*}{Predictor} & \multicolumn{4}{c}{Weighted L1 Loss} \\ 
{} &            ANX &             BIP &             DEP &             SUI \\
\midrule
\textsc{Mean}        &          .404 &          .489 &          .442 &         .506 \\
\textsc{Last}       &          .421 &         .513 &          .447 &         .542 \\
\textsc{Xgboost}    &          .415 &          .500 &          .452 &         .523 \\
\textsc{Model}      &  \textbf{.327} & \textbf{.401} & \textbf{.341} & \textbf{.432} \\
\textsc{Model-subreddit}     &     .412 & .491 & .423 & .473 \\
\bottomrule
\end{tabular}
\end{table*}

\noindent  \textbf{Quantitative results.} Table~\ref{tab:test_loss} shows, for each subreddit, the performance of the best model and the baselines on the test set, w.r.t. the Weighted L1 loss. Notably, \textsc{Mean} performs best among the baselines, outperforming even \textsc{XGB}. This advantage is easily explained by Figure~\ref{fig:heatmap_bipolar}: while both tend to collapse predictions around average, \textsc{Mean} does so less intensely, benefiting from lower error contributions on the extreme values, which carry more weight.
\textsc{Model}, in turn, shows a sizable error reduction w.r.t.\ \textsc{Mean}, which varies from 14.7\% on SUI to 22.8\% on DEP. We performed additional experiments to determine whether it is better to fit one model to each subreddit, indicated in Table~\ref{tab:test_loss} under the row \textsc{Model-subreddit}. The results show that the single model, trained on the entire data, performs better than the individual models.

\change{In Appendix~C, we perform experiments with two other subreddits, namely r/whatisthisthing and r/UnresolvedMysteries, to investigate if the model is not merely capturing a general trend in Reddit (e.g., an etiquette effect), but rather an improvement that is specific to mental health related subreddits. The results provide supporting evidence that predicted changes correspond to shifts in emotional tone.}

\noindent \textbf{Prediction examples.} We illustrate three cases that visually stand out in Figure~\ref{fig:heatmap_bipolar} where \textsc{Model} yields low, medium and high error, respectively. In the ANX plot, there is a short thread depicted by the closest point to the upper right corner, for which the emotional tone is $\textrm{EmT}(c_n) = 0.88$ and the prediction is $\hat y_\textsc{Model} = 0.74$. The author reports to have GAD (generalized anxiety disorder) and is seeking for help, but will be moving constantly during the next few years. The commenter congratulates the author for seeking help, acknowledging that it was one of the hardest steps for himself. Then he points out that many therapists do online sessions and suggests the author to look for someone specialized in anxiety disorders. In the final comment, the author acknowledges the difficulty in recognizing that \change{they} needed help and thanks the commenter for the kind words. In this case, the proposed model correctly captures the effect of the interaction. In the BIP plot, the leftmost point shows a thread for which $\textrm{EmT}(c_n) = -0.85$, but the model predicts $\hat y_\textsc{Model} = 0.01$. The author complains that he would like to be moving faster but is physically unable to do so. A handful of comments report similar feelings, to which the author says he can relate to. At the end, a commenter shares \change{their} arthritis issues, to which the author replies ``God dam* sounds irritating as hell''. In this specific case, we argue that the reply's tone may not be representative of the author's state (which is not clearly negative or positive). We investigated other discrepancies between the true and predicted values and found similar explanations. For instance, the point closest to the bottom right corner in \change{\SuicideWatch}\ has $\textrm{EmT}(c_n) = 0.88$, but the model predicts $\hat y_\textsc{Model} = -0.92$. The author says that \change{they are} being laid off and is experiencing suicidal ideation. A comment encourages their to meditate, but the author is too worried about their future finances. Another comment points out that there will always be another job. The author finishes the thread with two replies, explaining how tiring is the process of applying for another job, and that he is disappointed, on the edge and depressed, but no longer thinking about suicide and thanking the commenters. Although the author's status has improved, it is debatable whether the emotional tone associated with the last message (i.e., 0.88) accurately describes their state, since \change{they are} still very negative.


Next, in Table \ref{tab:depressionthread}, we take a deeper look at a \change{specific} thread from \change{\Depression}. Clearly, the encouragement provided by optimistic comments from other users leads to positive effects in the emotional tone of the thread author, corroborating our main findings from Figure \ref{fig:vader-hist2dCompPostComoUltimoComm}. Moreover, our model yields a very accurate prediction, $\hat y_\textsc{Model} = 0.51$, for the thread's author final $\textrm{EmT}(c_n) = 0.42$.
\begin{table}[t]
\begin{scriptsize}
\caption{\change{Example thread} extracted from \change{\Depression}. Our model accurately predicted the final emotional tone ($\hat y_\textsc{Model} = 0.51$). This example also illustrates the gradual improvement  of \change{the user's} emotional tone while receiving positive supporting from his mates. We can observe that, \change{by} the end, \change{their} emotional tone has significantly improved.} 
\centering
\begin{tabular}{|p{11.9cm}|c|}
\hline
\centering
\textbf{Post and Comments}&\textbf{EmT} \\
\hline
\textbf{Author}: Can somebody drag me out of this depression? Show me that there is still somebody who cares out there...& -0.14\\
\hline
\textbf{Commenter}: Hey, I care, I don't know you, I may never meet you, or hear you speak, but you matter, just as much as everyone else, stay strong [raised fist emoji] & 0.77 \\
\hline
\textbf{Author}: You have no idea how happy I am to hear this. You know, when someone is desperately in need of somebody else, these actions help them a lot! God bless you!&0.36  \\ 
\hline
\textbf{Commenter}: Any time, I know how it feels to be in need, I know it is hard, but you got this! [heart emoji] &-0.06\\
\hline
\textbf{Commenter}: You are a beautiful soul, don't worry you will be Ok, times will get better I promise. And I care!! :) I hope you smile and remember good people exist to take your hand through this. message me if you need to talk & 0.60\\
\hline
\textbf{Author}: You lighten me up! So great that somebody really care about those desperate words I wrote here! I wish all the best for you you beautiful soul! &0.42\\
\hline
\end{tabular}
\label{tab:depressionthread}
\end{scriptsize}
\end{table}


\begin{table*}[ht!]
\centering
\setlength{\tabcolsep}{6pt}
\caption{Performance on extreme test cases described in rows w.r.t.\ L1 loss \textsc{Model}, \textsc{Mean} and \textsc{Last} and \textsc{XGB} \change{(best shown in bold)}. Last three columns show percentage of the subset's threads in which \textsc{Model} outperforms each baseline \change{(italic indicates $< 50$\%)}.}
\label{tab:extreme_values}
\scriptsize
\begin{tabular}{l|rrrr|ccc}
\toprule
\multirow{3}{*}{Subset of test set} & \multicolumn{4}{c|}{\multirow{2}{*}{L1 Loss (unweighted)}} &  \multicolumn{3}{c}{\% Threads from subset in}\\
& & & & & \multicolumn{3}{c}{which \textsc{Model} outperforms} \\
  &  \textsc{Model} &   \textsc{Mean} &  \textsc{Last} &  \textsc{Xgb}  & \textsc{Mean}  &  \textsc{Last} & \textsc{XGB} \\
\midrule
$\Delta$EmT $>$ 95$^\textrm{th}$ perc.\ &     .522 &    .676 &    .521 &       \textbf{.499} &    71.1 &    50.8 &       51.4\\
$\Delta$EmT $>$ +1.0      &     .583 &    .743 &    .544 &       \textbf{.536} &    68.7 &           \textit{46.4} &              \textit{48.8} \\
$\textrm{EmT}(c_n)$ $>$ +0.8       &     \textbf{.469} &    .788 &    .668 &       .685 &    89.1 &    73.1 &       78.2 \\
$\Delta$EmT $<$  5$^\textrm{th}$ perc.\ &     \textbf{.426} &    .558 &    .563 &       .610 &    61.3 &    60.6 &       65.4 \\
$\Delta$EmT $<$ -1.0       &     \textbf{.440} &    .835 &    .806 &       .825 &    84.6 &    76.9 &       84.6 \\
$\textrm{EmT}(c_n)$ $<$ -0.8       &     \textbf{.403} &    .819 &    .887 &       .956 &    88.9 &    88.9 &       93.3 \\
\bottomrule
\end{tabular}
\end{table*}

\noindent \textbf{Extreme reactions.} We investigate whether \textsc{Model} outperforms the baselines for extreme cases, namely when (i) the absolute shift in emotional tone $|\Delta\textrm{EmT}| = |\textrm{EmT}(c_n) - \textrm{EmT}(c_\textrm{penultimate})|$ between the author's last and her second last comment is large and when (ii) $\textrm{EmT}(c_n)$ takes on extreme values. For this purpose, we compute statistics of the predictions for subsets of the 6\,311 threads in the test set. We consider as ``large shifts'' cases where
\begin{itemize}
    \item $\Delta\textrm{EmT} > +0.808$ (95$^\textrm{th}$ percentile),
    \item $\Delta\textrm{EmT} < -0.564$ (5$^\textrm{th}$ percentile),
    \item $\Delta\textrm{EmT} > +1.0$ \textbf{or}
    \item $\Delta\textrm{EmT} < -1.0$.
\end{itemize}
We consider \change{``extreme values'' as} cases where
\begin{itemize}
    \item $\textrm{EmT}(c_n) > +0.8$ \textbf{or}
    \item $\textrm{EmT}(c_n) < -0.8$.
\end{itemize}
This time we use the average absolute error (L1 Loss) since results are already conditioned on extreme values. Table~\ref{tab:extreme_values} shows the loss for \textsc{Model}, \textsc{Mean}, \textsc{Last} and \textsc{XGB}\change{, and} the percentage of threads from each subset such that \textsc{Model} outperforms each of the three baselines. 
In general, \textsc{Model} yields lower error than the baselines on all of the subsets, \change{other} than \textsc{Last} and \textsc{XGB} on extremely positive shifts (top two rows).  This is due to the fact that large improvements are often preceded by a very positive comment, although the converse is not necessarily true\change{; by} conditioning on high values of $\Delta\textrm{EmT}$, we artificially favor \textsc{Last} and \textsc{XGB}. On the other hand, since other users' comments tend to be positive in general, these baselines perform poorly on very negative shifts and on negative responses (three bottom rows). More importantly, considering the right panel of Table~\ref{tab:extreme_values}, \textsc{Model} outperforms \change{all} baselines when predicting extreme reactions in most threads\change{, except for the specific subset of threads where $\Delta\textrm{EmT} > +1.0$ (2$^\textrm{nd}$ row). Even in that subset,} the percentage is close to $50\%$ and the L1 loss difference is not large. 

\section{Discussion and Conclusions}
\label{sec:discussion}
In this work we consider a new, challenging task: to predict changes in the emotional tone (proxy for emotional state) of users after interactions in mental disorder online communities. The results from our prediction model indicated that, in spite of the inherent hardness of the task in hand, it is possible to predict with some accuracy the reaction of users to the interactions experienced in these platforms. The models we consider do not use \change{any user features, hence can be implemented without harming} their privacy and, as a byproduct, have the benefit of being applicable to users that have just joined a community.

A system capable of making accurate predictions can be used to assist users when composing comments, so as to increase the chance of making positive impacts on their readership. Possible ways to implement it are: as a service that runs in the backend of the online platform and returns the predicted emotional tone of the message's response while the user is composing a message, in real-time; or as a browser extension, in which case it wouldn't be limited to one specific website. \change{Such system can be used in conjunction with empathic rewriting techniques \cite{sharma2021www}.}

The fact that model training can take up to a minute may cause the impression that it is not suitable for real time applications. However, once the model is trained, it can process an entire thread including a comment that is being currently written in roughly 20ms.

There is room for improving the prediction model through the use of dilated RNNs, attention-based models and Transformers (this time for processing the sequence of comments). Particularly, attention models provide higher interpretability that can be used to pinpoint which comments were the most important for a given prediction. Another promising class of models is the Graph Neural Network, which could be used to process the content present in discussion trees along with their structure, instead of treating them as sequential inputs. We leave these investigations for future work.

We also showed that \change{data from other mental health-related subreddits can improve predictions for a target subreddit. It} is possible that other subreddits can be included for training. More data can also benefit the training of more complex models.  

\change{Finally}, we underline that deep learning models have been successfully used to learn good latent representations (encodings) of the input data by training in one task and then using these encodings in another task \cite{socher2017tacl,radford2018improving}. While most models for predicting mental health disorders from online data focus on non-temporal, textual data exclusively from the subject being predicted, our model attempts to capture the expected responses from a sequence of interactions. Significant deviations from the expected trajectory could be correlated with \change{more severe illnesses} and, hence, potentially useful for detection tasks. Moreover, the latent state sequences associated with each user can be used to obtain encodings. These temporal encodings can, in turn, be clustered so as to better understand the existing types of users, how they respond to interactions and to investigate which types of intervention work best for them.


\section*{Acknowledgments}

This study was partially suported by grants from Coordenação de Aperfeiçoamento de Pessoal de Nível Superior – Brasil (CAPES), Conselho Nacional de Desenvolvimento Científico e Tecnológico (CNPq) and Fundação de Amparo à Pesquisa de Minas Gerais (FAPEMIG). We also thank Shiri Dori-Hacohen for her careful review and insightful comments.




 





\bibliographystyle{elsarticle-num}
\bibliography{references}

\begin{thebibliography}{10}
\expandafter\ifx\csname url\endcsname\relax
  \def\url#1{\texttt{#1}}\fi
\expandafter\ifx\csname urlprefix\endcsname\relax\def\urlprefix{URL }\fi
\expandafter\ifx\csname href\endcsname\relax
  \def\href#1#2{#2} \def\path#1{#1}\fi

\bibitem{silveira2021fgcs}
B.~Silveira, H.~S. Silva, F.~Murai, A.~P.~C. {da Silva}, Predicting user
  emotional tone in mental disorder online communities, Future Generation
  Computer Systems 125 (2021) 641--651.
\newblock \href {https://doi.org/https://doi.org/10.1016/j.future.2021.07.014}
  {\path{doi:https://doi.org/10.1016/j.future.2021.07.014}}.

\bibitem{world2013mental}
WHO, Mental health action plan 2013-2020 (2013).

\bibitem{world2017depression}
WHO, Depression and other common mental disorders: global health estimates,
  Tech. rep., World Health Organization (2017).

\bibitem{barney2006stigma}
L.~J. Barney, K.~M. Griffiths, A.~F. Jorm, H.~Christensen, Stigma about
  depression and its impact on help-seeking intentions, Australian \& New
  Zealand Journal of Psychiatry 40~(1) (2006) 51--54.

\bibitem{10.1371/journal.pone.0215476}
R.~M. Merchant, D.~A. Asch, P.~Crutchley, L.~H. Ungar, S.~C. Guntuku, J.~C.
  Eichstaedt, S.~Hill, K.~Padrez, R.~J. Smith, H.~A. Schwartz, Evaluating the
  predictability of medical conditions from social media posts, PLOS ONE 14~(6)
  (2019) 1--12.
\newblock \href {https://doi.org/10.1371/journal.pone.0215476}
  {\path{doi:10.1371/journal.pone.0215476}}.

\bibitem{facebook2018}
C.~Card, How facebook ai helps suicide prevention, Facebook Newsroom (2018).

\bibitem{mind2020}
A.~Zaman, V.~Silenzio, H.~Kautz, Mind: A tool for mental health screening and
  support of therapy to improve clinical and research outcomes, in: EAI
  International Conference on Pervasive Computing Technologies for Healthcare,
  PervasiveHealth '20, ACM, New York, NY, USA, 2020, p. 423–426.
\newblock \href {https://doi.org/10.1145/3421937.3421959}
  {\path{doi:10.1145/3421937.3421959}}.

\bibitem{munmun2020}
S.~Chancellor, M.~De~Choudhury, Methods in predictive techniques for mental
  health status on social media: a critical review, Digital Medicine 3~(1)
  (2020) 43.

\bibitem{Choudhury2014MentalHD}
M.~De~Choudhury, S.~De, Mental health discourse on reddit: Self-disclosure,
  social support, and anonymity, in: International Conference on Web and Social
  Media, 2014.

\bibitem{gkotsis2016language}
G.~Gkotsis, A.~Oellrich, T.~Hubbard, R.~Dobson, M.~Liakata, S.~Velupillai,
  R.~Dutta, The language of mental health problems in social media, in:
  Workshop on Computational Lingusitics and Clinical Psychology, 2016.

\bibitem{gkotsis2017characterisation}
G.~Gkotsis, A.~Oellrich, S.~Velupillai, M.~Liakata, T.~J. Hubbard, R.~J.
  Dobson, R.~Dutta, Characterisation of mental health conditions in social
  media using informed deep learning, Scientific reports 7 (2017) 45141.

\bibitem{park2018harnessing}
A.~Park, M.~Conway, Harnessing reddit to understand the written-communication
  challenges experienced by individuals with mental health disorders: Analysis
  of texts from mental health communities, Journal of Medical Internet Research
  20~(4) (2018).

\bibitem{thorstad2019}
R.~Thorstad, P.~Wolff, Predicting future mental illness from social media: A
  big-data approach, Behavior Research Methods 51~(4) (2019) 1586--1600.
\newblock \href {https://doi.org/10.3758/s13428-019-01235-z}
  {\path{doi:10.3758/s13428-019-01235-z}}.

\bibitem{Saha2020.08.07.20170548}
K.~Saha, J.~Torous, E.~D. Caine, M.~De~Choudhury, Social media reveals
  psychosocial effects of the covid-19 pandemic, medRxiv (2020).
\newblock \href {https://doi.org/10.1101/2020.08.07.20170548}
  {\path{doi:10.1101/2020.08.07.20170548}}.

\bibitem{shen2017detecting}
J.~H. Shen, F.~Rudzicz, Detecting anxiety on reddit, in: Workshop on
  Computational Linguistics and Clinical Psychology—From Linguistic Signal to
  Clinical Reality, 2017, pp. 58--65.

\bibitem{yusof2018assessing}
N.~F.~A. Yusof, C.~Lin, F.~Guerin, Assessing the effectiveness of affective
  lexicons for depression classification, in: International Conference on
  Applications of Natural Language to Information Systems, Springer, 2018, pp.
  65--69.

\bibitem{murrieta2018depression}
J.~Murrieta, C.~C. Frye, L.~Sun, L.~G. Ly, C.~S. Cochancela, E.~V. Eikey, \#
  depression: Findings from a literature review of 10 years of social media and
  depression research, in: International Conference on Information, Springer,
  2018, pp. 47--56.

\bibitem{lee2018advanced}
K.~S. Lee, H.~Lee, W.~Myung, G.-Y. Song, K.~Lee, H.~Kim, B.~J. Carroll, D.~K.
  Kim, Advanced daily prediction model for national suicide numbers with social
  media data, Psychiatry investigation 15~(4) (2018) 344.

\bibitem{Gaur2018}
M.~Gaur, U.~Kursuncu, A.~Alambo, A.~Sheth, R.~Daniulaityte, K.~Thirunarayan,
  J.~Pathak, "let me tell you about your mental health!": Contextualized
  classification of reddit posts to dsm-5 for web-based intervention, in: ACM
  International Conference on Information and Knowledge Management, 2018, p.
  753–762.

\bibitem{coppersmith2018natural}
G.~Coppersmith, R.~Leary, P.~Crutchley, A.~Fine, Natural language processing of
  social media as screening for suicide risk, Biomedical informatics insights
  10 (2018).

\bibitem{wongkoblap2019modeling}
A.~Wongkoblap, M.~A. Vadillo, V.~Curcin, Modeling depression symptoms from
  social network data through multiple instance learning, AMIA Summits on
  Translational Science Proceedings 2019 (2019) 44.

\bibitem{fraga2018wi}
B.~{Silveira Fraga}, A.~P. {Couto da Silva}, F.~{Murai}, Online social networks
  in health care: A study of mental disorders on reddit, in: 2018 IEEE/WIC/ACM
  International Conference on Web Intelligence (WI), 2018, pp. 568--573.
\newblock \href {https://doi.org/10.1109/WI.2018.00-36}
  {\path{doi:10.1109/WI.2018.00-36}}.

\bibitem{yates2017depression}
A.~Yates, A.~Cohan, N.~Goharian, Depression and self-harm risk assessment in
  online forums, arXiv preprint arXiv:1709.01848 (2017).

\bibitem{Althoff2016NaturalLP}
T.~Althoff, K.~Clark, J.~Leskovec, Large-scale analysis of counseling
  conversations: An application of natural language processing to mental
  health, Transactions of the Association for Computational Linguistics 4
  (2016) 463.

\bibitem{chen2018mood}
X.~Chen, M.~D. Sykora, T.~W. Jackson, S.~Elayan, What about mood swings:
  Identifying depression on twitter with temporal measures of emotions, in:
  Companion Proceedings of the The Web Conference 2018, 2018, pp. 1653--1660.
\newblock \href {https://doi.org/10.1145/3184558.3191624}
  {\path{doi:10.1145/3184558.3191624}}.

\bibitem{wolohan-etal-2018-detecting}
J.~Wolohan, M.~Hiraga, A.~Mukherjee, Z.~A. Sayyed, M.~Millard, Detecting
  linguistic traces of depression in topic-restricted text: Attending to
  self-stigmatized depression with {NLP}, in: International Workshop on
  Language Cognition and Computational Models, 2018, pp. 11--21.

\bibitem{dutta2018measuring}
S.~Dutta, J.~Ma, M.~De~Choudhury, Measuring the impact of anxiety on online
  social interactions, in: International AAAI Conference on Web and Social
  Media, 2018.

\bibitem{gruda2019feeling}
D.~Gruda, S.~Hasan, Feeling anxious? perceiving anxiety in tweets using machine
  learning, Computers in Human Behavior 98 (2019) 245--255.

\bibitem{sahota2019bipolar}
P.~K. Sahota, P.~L. Sankar, Bipolar disorder, genetic risk, and reproductive
  decision-making: A qualitative study of social media discussion boards,
  Qualitative health research (2019).

\bibitem{baba2019detecting}
T.~Baba, K.~Baba, D.~Ikeda, Detecting mental health illness using short
  comments, in: International Conference on Advanced Information Networking and
  Applications, Springer, 2019, pp. 265--271.

\bibitem{shing-etal-2018-expert}
H.-C. Shing, S.~Nair, A.~Zirikly, M.~Friedenberg, H.~Daum{\'e}~III, P.~Resnik,
  Expert, crowdsourced, and machine assessment of suicide risk via online
  postings, in: Workshop on Computational Linguistics and Clinical Psychology:
  From Keyboard to Clinic, Association for Computational Linguistics, 2018, pp.
  25--36.

\bibitem{10.1145/3134727}
K.~Saha, M.~De~Choudhury, Modeling stress with social media around incidents of
  gun violence on college campuses, Proc. ACM Hum.-Comput. Interact. 1~(CSCW)
  (Dec. 2017).

\bibitem{pirina-coltekin-2018-identifying}
I.~Pirina, {\c{C}}.~{\c{C}}{\"o}ltekin, Identifying depression on {R}eddit: The
  effect of training data, in: {EMNLP} Workshop {SMM}4{H}: The 3rd Social Media
  Mining for Health Applications Workshop {\&} Shared Task, Association for
  Computational Linguistics, 2018, pp. 9--12.
\newblock \href {https://doi.org/10.18653/v1/W18-5903}
  {\path{doi:10.18653/v1/W18-5903}}.

\bibitem{info:doi/10.2196/jmir.9840}
A.~E. Alada{\u{g}}, S.~Muderrisoglu, N.~B. Akbas, O.~Zahmacioglu, H.~O. Bingol,
  Detecting suicidal ideation on forums: Proof-of-concept study, J Med Internet
  Res 20~(6) (2018) e215.

\bibitem{10.1145/3173574.3174240}
S.~Chancellor, A.~Hu, M.~De~Choudhury, Norms matter: Contrasting social support
  around behavior change in online weight loss communities, in: SIGCHI
  Conference on Human Factors in Computing Systems, Association for Computing
  Machinery, 2018, p. 1–14.

\bibitem{ireland-iserman-2018-within}
M.~Ireland, M.~Iserman, Within and between-person differences in language used
  across anxiety support and neutral {R}eddit communities, in: Workshop on
  Computational Linguistics and Clinical Psychology: From Keyboard to Clinic,
  Association for Computational Linguistics, 2018, pp. 182--193.

\bibitem{de2016discovering}
M.~De~Choudhury, E.~Kiciman, M.~Dredze, G.~Coppersmith, M.~Kumar, Discovering
  shifts to suicidal ideation from mental health content in social media, in:
  SIGCHI Conference on Human Factors in Computing Systems, ACM, 2016, pp.
  2098--2110.

\bibitem{gkotsis2017}
G.~Gkotsis, A.~Oellrich, S.~Velupillai, M.~Liakata, T.~J.~P. Hubbard, R.~J.~B.
  Dobson, R.~Dutta, Characterisation of mental health conditions in social
  media using informed deep learning, Scientific Reports 7~(1) (2017) 45141.
\newblock \href {https://doi.org/10.1038/srep45141}
  {\path{doi:10.1038/srep45141}}.

\bibitem{10.1145/3159652.3159725}
F.~Sadeque, D.~Xu, S.~Bethard, Measuring the latency of depression detection in
  social media, in: ACM International Conference on Web Search and Data Mining,
  WSDM '18, ACM, 2018, p. 495–503.
\newblock \href {https://doi.org/10.1145/3159652.3159725}
  {\path{doi:10.1145/3159652.3159725}}.

\bibitem{ive-etal-2018-hierarchical}
J.~Ive, G.~Gkotsis, R.~Dutta, R.~Stewart, S.~Velupillai, Hierarchical neural
  model with attention mechanisms for the classification of social media text
  related to mental health, in: Workshop on Computational Linguistics and
  Clinical Psychology: From Keyboard to Clinic, Association for Computational
  Linguistics, 2018, pp. 69--77.
\newblock \href {https://doi.org/10.18653/v1/W18-0607}
  {\path{doi:10.18653/v1/W18-0607}}.

\bibitem{coppersmith-etal-2014-quantifying}
G.~Coppersmith, M.~Dredze, C.~Harman, Quantifying mental health signals in
  {T}witter, in: Workshop on Computational Linguistics and Clinical Psychology:
  From Linguistic Signal to Clinical Reality, 2014, pp. 51--60.

\bibitem{edwards2017}
T.~Edwards, N.~S. Holtzman, A meta-analysis of correlations between depression
  and first person singular pronoun use, Journal of Research in Personality 68
  (2017) 63--68.
\newblock \href {https://doi.org/https://doi.org/10.1016/j.jrp.2017.02.005}
  {\path{doi:https://doi.org/10.1016/j.jrp.2017.02.005}}.

\bibitem{loveys-etal-2017-small}
K.~Loveys, P.~Crutchley, E.~Wyatt, G.~Coppersmith, Small but mighty: Affective
  micropatterns for quantifying mental health from social media language, in:
  Workshop on Computational Linguistics and Clinical Psychology {---} From
  Linguistic Signal to Clinical Reality, Association for Computational
  Linguistics, 2017, pp. 85--95.

\bibitem{mauss2009}
I.~B. Mauss, M.~D. Robinson, Measures of emotion: A review, Cognition \&
  emotion 23~(2) (2009) 209--237.
\newblock \href {https://doi.org/10.1080/02699930802204677}
  {\path{doi:10.1080/02699930802204677}}.

\bibitem{kayla2018}
K.~N. Jordan, J.~W. Pennebaker, C.~Ehrig, The 2016 u.s. presidential candidates
  and how people tweeted about them, SAGE Open 8~(3) (2018) 2158244018791218.

\bibitem{chancellor2016}
S.~Chancellor, T.~Mitra, M.~De~Choudhury, Recovery amid pro-anorexia: Analysis
  of recovery in social media, SIGCHI Conference on Human Factors in Computing
  Systems (2016) 2111--2123.

\bibitem{DBLP:conf/cscw/ChoudhuryCHH14}
M.~De~Choudhury, S.~Counts, E.~J. Horvitz, A.~Hoff, Characterizing and
  predicting postpartum depression from shared facebook data, in: ACM
  Conference on Computer Supported Cooperative Work, CSCW '14, ACM, 2014, p.
  626–638.
\newblock \href {https://doi.org/10.1145/2531602.2531675}
  {\path{doi:10.1145/2531602.2531675}}.

\bibitem{hutto2014vader}
C.~J. Hutto, E.~Gilbert, Vader: A parsimonious rule-based model for sentiment
  analysis of social media text, in: International AAAI conference on weblogs
  and social media, 2014.

\bibitem{sanh2020distilbert}
V.~Sanh, L.~Debut, J.~Chaumond, T.~Wolf, Distilbert, a distilled version of
  bert: smaller, faster, cheaper and lighter (2020).
\newblock \href {http://arxiv.org/abs/1910.01108} {\path{arXiv:1910.01108}}.

\bibitem{chung2014arxiv}
J.~Chung, C.~Gulcehre, K.~Cho, Y.~Bengio, Empirical evaluation of gated
  recurrent neural networks on sequence modeling (2014).
\newblock \href {http://arxiv.org/abs/1412.3555} {\path{arXiv:1412.3555}}.

\bibitem{jozefowicz2015pmlr}
R.~Jozefowicz, W.~Zaremba, I.~Sutskever, An empirical exploration of recurrent
  network architectures, in: F.~Bach, D.~Blei (Eds.), International Conference
  on Machine Learning, Vol.~37 of Proceedings of Machine Learning Research,
  PMLR, Lille, France, 2015, pp. 2342--2350.

\bibitem{pennebaker01}
J.~W. Pennebaker, M.~E. Francis, R.~J. Booth, Linguistic Inquiry and Word
  Count, Lawerence Erlbaum Associates, Mahwah, NJ, 2001.

\bibitem{bert2018}
J.~Devlin, M.-W. Chang, K.~Lee, K.~Toutanova, Bert: Pre-training of deep
  bidirectional transformers for language understanding (2019).
\newblock \href {http://arxiv.org/abs/1810.04805} {\path{arXiv:1810.04805}}.

\bibitem{xgboost}
T.~Chen, C.~Guestrin, {XGBoost}: A scalable tree boosting system, in: ACM
  SIGKDD International Conference on Knowledge Discovery and Data Mining, KDD
  '16, ACM, New York, NY, USA, 2016, pp. 785--794.
\newblock \href {https://doi.org/10.1145/2939672.2939785}
  {\path{doi:10.1145/2939672.2939785}}.

\bibitem{sharma2021www}
A.~Sharma, I.~W. Lin, A.~S. Miner, D.~C. Atkins, T.~Althoff, Towards
  facilitating empathic conversations in online mental health support: A
  reinforcement learning approach, WWW '21, Association for Computing
  Machinery, New York, NY, USA, 2021, p. 194–205.
\newblock \href {https://doi.org/10.1145/3442381.3450097}
  {\path{doi:10.1145/3442381.3450097}}.

\bibitem{socher2017tacl}
R.~Socher, A.~Karpathy, Q.~Le, C.~Manning, A.~Ng, Grounded compositional
  semantics for finding and describing images with sentences, Transactions of
  the Association for Computational Linguistics 2~(0) (2014) 207--218.

\bibitem{radford2018improving}
A.~Radford, K.~Narasimhan, T.~Salimans, I.~Sutskever,
  \href{https://s3-us-west-2.amazonaws.com/openai-assets/research-covers/language-unsupervised/language_understanding_paper.pdf}{Improving
  language understanding by generative pre-training} (2018).
\newline\urlprefix\url{https://s3-us-west-2.amazonaws.com/openai-assets/research-covers/language-unsupervised/language_understanding_paper.pdf}

\end{thebibliography}

\appendix
\section{\change{Hyperparameter Tuning and Experimental Details}}\label{sec:tuning}

Tables~\ref{tab:tuning_rnn} and \ref{tab:tuning_xgb} list the hyperparameters values considered in the grid search, for the proposed model and XGBoost, respectively. We also include the best hyperparameter configuration in each case, i.e.\ the one that minimizes the objective function on the validation set, which consists of 10\% of the entire data. Our python implementation uses the pytorch\footnote{https://pytorch.org} and xgboost\footnote{https://xgboost.readthedocs.io} libraries.
\setcounter{table}{0}

\begin{table*}[ht!]
\centering
\scriptsize
\caption{\textsc{Model}'s hyperparameter values tested in grid search and values in best configuration. Note$^*$: dropout should not be used in the last layer and, thus, is always set to 0.0 when the number of layers is 1.}
\label{tab:tuning_rnn}
\begin{tabular}{lcc}
\toprule
Hyperparameter & values & best config.\\
\midrule
hidden size & 4, 16, 64 & 64 \\
num layers & 1, 2 & 2 \\
dropout$^*$ & 0.0, 0.1, 0.2, 0.5 & 0.0 \\
num directions & 1, 2 & 1 \\
\bottomrule
\end{tabular}
\end{table*}

\begin{table*}[ht!]
\centering
\scriptsize
\caption{\textsc{XGBoost}'s hyperparameter values tested in grid search and values in best configuration. Note $^*$: when the maximum number of estimators was set to 500, the fit continued until that number of trees were generated, although it could have finished earlier due to the early stopping mechanism.}
\label{tab:tuning_xgb}
\begin{tabular}{lcc}
\toprule
Hyperparameter & values & best config.\\
\midrule
learning rate & $10^{-3}$, $10^{-2}$, $10^{-1}$ & $10^{-2}$ \\
max depth & 1, 3, 5 & 5 \\
min child weight & 1, 3, 5 & 1\\
subsample & 0.5, 0.7 & 0.7 \\
colsample by tree & 0.5, 0.7 & 0.7  \\
n estimators & 100, 200, 500 & 500$^*$ \\
objective & MSE, Weighted L1 Loss & MSE \\
\bottomrule
\end{tabular}
\end{table*}

\change{The experiments we perform have three main sources of uncertainty:
\begin{itemize}
    \item the data split into training, validation and test sets;
    \item the initial weights of the neural network \textsc{Model};
    \item the initial value of \textsc{XGB}'s random state.
\end{itemize}
}

\change{We reduce the variance of the results by controlling some of these sources of uncertainty. Specifically, the seeds for the random number generators are set before:
\begin{itemize}
    \item splitting the training, validation and test sets;
    \item instantiating \textsc{Model} and \textsc{XGB}.
\end{itemize}
Note that the data split is done only once, whereas \textsc{Model} and \textsc{XGB} are instantiated once for every configuration in the hyperparameter tuning. 
}

\change{We conducted some experiments considering different splits of the data and different model initializations and the results did not change significantly, likely due to the large number of instances used for training and testing.}


\section{Results for other loss functions}\label{sec:other_losses}

Table~\ref{tab:other_losses} shows the L1 and MSE losses for the baselines and for the proposed model. These were omitted from the main text because they fail to capture the quality of the results. Specifically, note that although \textsc{Xgboost-MSE} yields the lowest L1 and MSE losses, it does so by collapsing predictions around the average, even more intensely than the second best baseline w.r.t.\ these metrics, \textsc{Mean} (please refer to Figure~\ref{fig:heatmap_bipolar}).
\setcounter{table}{0}
\begin{table*}[ht!]
\setlength{\tabcolsep}{6pt}
\centering
\caption{Results w.r.t.\ L1 and MSE losses. \textsc{Model} is trained on entire 2017 dataset, whereas \textsc{Model-Subreddit} is trained on target subreddit (column). } 
\label{tab:other_losses}
\scriptsize
\begin{tabular}{l|rrrr|rrrr}
\toprule
\multirow{2}{*}{Predictor}  & \multicolumn{4}{c|}{L1 Loss} & \multicolumn{4}{c}{MSE Loss} \\ 
{} &                       ANX &             BIP &             DEP &             SUI &                      ANX &             BIP &             DEP &             SUI \\
\midrule
\textsc{Unchanged}  &          .362 &          .306 &          .365 &          .347&          .212 &          .154 &          .216 &          .199 \\
\textsc{Mean}     &          .279 &          .241 &          .286 &          .276 &          .125  &          .101 &          .131 &          .127 \\
\textsc{Last}      &          .299 &          .295 &          .312 &          .330  &          .148 &          .146 &          .162 &          .181 \\
\textsc{Xgboost}   & .236 & .231 & .256 & .265 & .091 & .088 & .106 & .115 \\
\textsc{Model}     &          .281 &          .276 &          .312 &          .340  &          .125 &          .128 &          .156 &          .185 \\
\textsc{Model-Subreddit}     &          .334 &          .254 &          .318 &          .302 &                  .182 &          .106 &          .159 &          .151 \\
\bottomrule
\end{tabular}
\end{table*}

\section{Experiments on subreddits unrelated to mental health}\label{sec:othersubs}


\change{We perform additional experiments to investigate if the model is not merely capturing a general trend in Reddit (e.g., an etiquette effect), but rather an improvement that is specific to mental health related subreddits. More precisely, we analyze two different subreddits: r/whatisthisthing and r/UnresolvedMysteries. whatisthisthing follows a completely different dynamic to that seen on the original mental health subreddits, in which discussion threads tend to be longer, i.e., where the O.P.\ interacts multiple times with the commenters. UnresolvedMysteries, in turn, is taken as an example of subreddit that follows a similar dynamic to that of the original subreddits, but where major changes in the O.P.'s emotional tone are not expected, since the subreddit mainly revolves around discussing mysteries and other topics that are generally not related to an individual's emotional aspects. We show that the proposed model has characteristics that make it suitable specifically for online mental health communities and, therefore, may not work on subreddits that follow different dynamics.}

\change{We crawled the 2017 data for each subreddit and applied the same filtering procedures to select the threads used in the original subreddits analyzed. We found that: out of the 53720 (resp.\ 1237) posts gathered from whatisthisthing (resp.\ UnresolvedMysteries), 2.3\% (resp.\ 44\%) passed our filters. This underlines the aforementioned differences and similarities between these two subreddits and the original analyzed ones. We then proceeded to follow two alternative approaches: (i) use each new subreddit as a test set for our already trained model or (ii) train a new model using each new subreddit as a training set. Approach (ii) yielded worst results, likely due to the small number of samples on each dataset (respectively, 1237 and 3416 posts, compared to 63117 posts for the original model).}

\change{In these two subreddits, under the two different experimental setups, the proposed model is outperformed by at least one of the baselines. Results are shown in Table~\ref{tab:control_subs}. This can be seen as evidence for the case that our model is specifically suited for mental health related online forums, where there is a clear dynamic regarding the emotional tone of users on interactive threads. In sum, we believe that the initial results regarding the number of filtered posts from each subreddit, coupled with the fact that the model was outperformed by at least one of the baselines on all experimental setups, provide evidence that, in the context of mental health online forums, the measured shifts do correspond to a good approximation of the user’s emotional tone. While there are some shortcomings regarding the accuracy of sentiment analysis methods, we show that, given the available tools for extracting sentiment score, it is possible to build a model that accurately predicts shifts in the user’s emotional tone.}

\setcounter{table}{0}
\begin{table}
\centering
\caption{Results w.r.t.\ the Weighted L1 Loss when predicting on control subreddits \textsc{r/UnresolvedMysteries} (UM) and \textsc{r/WhatIsThisThing} (WTT). Columns 2-3 correspond to \textsc{Model} trained on data from a specific subreddit. Columns 4-5 correspond to \textsc{Model} trained on the four original subreddits. \textsc{Model} is outperfomed by \textsc{Last} when trained with UM data and by all other predictors when trained with the WTT data. The original \textsc{Model} is also outperformed when tested on the two control subreddits. }
  \label{tab:control_subs}
\scriptsize
  \begin{tabular}{l|cc|cc}
    \toprule
    \multirow{3}{*}{Predictor} &
      \multicolumn{2}{c|}{\textsc{Trained on Sub}} &
      \multicolumn{2}{c}{\textsc{Original Model}}  \\ \cmidrule{2-5}\cmidrule{2-5}
      & \multicolumn{2}{c|}{\textsc{Subreddit}} &  \multicolumn{2}{c}{\textsc{Subreddit}}  \\
    & ~~UM~~ & WTT & ~~UM~~ & WTT  \\
    \midrule
    UNCHANGED & .731  & .487  & .472 & .431 \\
    MEAN & .699 & .478 & \textbf{.455} & .392  \\
    LAST & \textbf{.633} & \textbf{.439} & .475 & \textbf{.375}  \\
    MODEL & .688 & .507 & .476 & .382  \\
    \bottomrule
  \end{tabular}
\end{table}





\end{document}